\documentclass[11pt]{article}

\newtheorem{theorem}{Theorem}

\newtheorem{corollary}{Corollary}
\newenvironment{proof}{\begin{list}{$\!\!${\bf Proof} \rule{1pt}{0pt}}
{\setlength{\leftmargin}{0pt}\setlength{\itemindent}{30pt}\setlength{\listparindent}
{15pt}}\item}{\rule{0.3em}{0mm}\hfill\framebox[1.2ex]{\rule{0.3em}{0mm}}
\end{list}}

\usepackage{amsmath,amssymb, comment}

\newcommand{\LL}{{\mathcal L}}
\newcommand{\Lcal}{\mathcal{L}}

\newcommand{\z}{\mathbf{z}}
\newcommand{\mb}{\mathbf{m}}
\newcommand{\rb}{\mathbf{r}}
\newcommand{\p}{\mathbf{p}}

\newcommand{\R}{ \mathbb{R}}
\newcommand{\sgn}{{\rm sign }}
\newcommand{\diag}{{\rm diag}}

\usepackage[colorlinks=true, urlcolor=blue, linkcolor=blue, citecolor=red]{hyperref}

\begin{document}

\title{Preconditioning for Accelerated Gradient Descent Optimization and Regularization}

\author{Qiang Ye\thanks{Department of Mathematics, University of Kentucky,
Lexington, KY 40506. {\tt qye3@uky.edu}. Research supported in part by NSF under grants DMS-2208314, IIS-2327113,  and ITE-2433190.
}
}
\date{}

\maketitle

\begin{abstract}
Accelerated training algorithms, such as adaptive learning rates (or preconditioning) and various normalization methods, are widely used but not fully understood. When regularization is introduced, standard optimizers like adaptive learning rates may not perform effectively. This raises the need for alternative regularization approaches such as AdamW and the question of how to properly combine regularization with preconditioning. In this paper, we address these challenges using the theory of preconditioning as follows: (1) We explain how AdaGrad, RMSProp, and Adam accelerates training through improving Hessian conditioning; (2) We explore the interaction between $L_2$-regularization and preconditioning, demonstrating that AdamW \cite{loshchilov2018decoupled} amounts to selecting the underlying intrinsic parameters for regularization, and we derive a generalization for the $L_1$-regularization;  and (3) We demonstrate how various normalization methods such as input data normalization, batch normalization, and layer normalization accelerate training by improving Hessian conditioning.
Our analysis offers a unified mathematical framework for understanding various acceleration techniques or deriving appropriate regularization schemes.
\end{abstract}


\section{Introduction}

Accelerated gradient descent algorithms such as AdaGrad, RMSProp, and Adam (see \cite{duchi2011adaptive,RMSProp,2015-kingma})
have played a pivotal role in the success of deep learning. However, their underlying mechanisms are not fully understood. These methods are widely interpreted as adaptive learning rate methods where the learning rate for each individual parameter adapts according to the magnitude of the partial derivative with respect to that parameter so that a smaller learning rate is used for a parameter with a larger derivative; see \cite[Sec. 8.5]{Goodfellow-et-al-2016} and \cite[Sec. 7.3.3]{bishop2023learning}. One limitation of this interpretation is that an appropriate adaptive learning rate can simply be chosen to be inversely proportional to the absolute value of the partial derivative at each iteration, which would render the gradient descent update to use the signs of the partial derivatives only, eliminating all other gradient information. On the other hand, these algorithms are also frequently considered as preconditioning methods \cite[Sec. 8.4.6]{murphy2013machine}, though it remains unclear why the particular diagonal preconditioners constructed by these algorithms from the gradients rather than the Hessian are effective.

Of more practical importance is how we use adaptive learning rate in combination with regularization. \cite{loshchilov2018decoupled} points out that directly applying an adaptive learning rate method to a $L_2$ regularized loss leads to an algorithm  that differ from weight decay. They further show that the $L_2$ regularization is not effective with Adam and advocate the use of AdamW, which decouples weight decay from the adaptive learning rate. This discrepancy between the $L_2$ regularization and weight decay in the adaptive learning rate (or preconditioning) setting raises an important question  about how to properly combine adaptive learning rates with regularization. This is important as the answer is even less clear for other regularization methods such as the $L_1$ regularization or gradient regularizations (see \cite{barrett2020implicit,karakida2023understanding,smith2021on,zhao2022penalizing}).

Normalization methods such as input data normalization and batch normalization \cite{Ioffe15} are also critical methods in training neural network models. Intuitively, normalizing each input or hidden variable component across the dataset to have  similar magnitudes prevents the situation that, when an input/variable is much larger than others, a small change in the corresponding weight causes a disproportionately  large change in the output  \cite[Sec. 7.4]{bishop2023learning}. While this improves the stability of the network with respect to trainable parameters, the exact benefits of normalization in improving optimization are not fully understood. Yet,  other important normalization methods such as LayerNorm of \cite{LN} and GroupNorm of \cite{wu2018group}, 
which normalize a hidden variable in a dimension other than the data/batch dimension, have also been found to be effective in training. In general, there is a lack of understanding as to why these different varieties of normalization methods seem to work similarly.

In this paper, we address these challenging questions from the theory of preconditioning as follows: (1) We explain how AdaGrad, RMSProp, and Adam accelerates training through improving Hessian conditioning; (2) We explore the interaction between $L_2$-regularization and preconditioning, demonstrating that AdamW \cite{loshchilov2018decoupled} amounts to selecting underlying intrinsic parameters for regularization, and we derive a generalization for the $L_1$-regularization;
and (3) We demonstrate how normalization methods like input data normalization, batch normalization, and layer normalization accelerate training by improving Hessian conditioning, which reveals some underlying connections of these methods. Our analysis offers a unified mathematical framework for understanding various acceleration techniques and deriving appropriate regularization schemes.

Throughout, multiplications or divisions of vectors of the same dimension such as $u v, 1/v$  or functions of vectors such as $ v^2$ are entrywise. $\|\cdot\|$  denotes the 2-norm unless specified otherwise. $\kappa (A)$ denotes the condition number of a matrix $A$ in 2-norm. $e$ denotes a vector of ones of appropriate dimension, i.e. $e=[1, 1, \cdots, 1]^T$.

\section{Theory for Preconditioned Gradient Descent}
\label{pgd_section}
Given  a loss function $\mathcal{L}(\p): \mathbb{R}^{n } \rightarrow \mathbb{R}$, the gradient descent (GD) method updates an approximate minimizer $\p_t$, starting from an initial approximation $\p_0$,  as:
\begin{equation}\label{eq:gd}
     \p_{t+1} = \p_t - \alpha \nabla \mathcal{L} (\p_t),   \;\; \mbox{ for } t=0,1,2 \cdots
\end{equation}
where $\alpha$ is a learning rate. We use  $\lambda_{\text{min}} (A)$ and $\lambda_{\text{max}} (A)$ (or simply $\lambda_{\text{min}}$ and $\lambda_{\text{max}} $)  to denote respectively the minimum and the maximum eigenvalues of a square matrix $A$, and use  $ \kappa \left(A\right) :=\frac{\lambda_{\text{max}} \left(A\right)}{\lambda_{\text{min}} \left(A\right)}$ to
denote the spectral condition number of a symmetric positive definite $A$. A general local convergence result  to a local minimizer  describes the asymptotic convergence rate in terms of the condition number of Hessian.

\begin{theorem}
Assume $\mathcal{L}(\p): \mathbb{R}^{n } \rightarrow \mathbb{R}$ is twice continuously differentiable and  $\p^{*}$ is such that $\nabla \mathcal{L}(\p^*)=0$ and the Hessian matrix $\nabla^2 \mathcal{L}(\p^*)$ is positive definite. Then for any $\epsilon>0$, there is a  small neighborhood around $\p^{*}$ such that, for any initial approximation $\p_0$ in that neighborhood, the GD iterations (\ref{eq:gd}) satisfy
 \begin{equation}\label{eq:oneiter}
 \| \p_{t+1} - \p^{*} \| \leq (r+\epsilon)  \|\p_t - \p^{*} \|,
\end{equation}
where
\[
r = \max \{ \left|1 - \alpha \lambda_{\text{min}} \right|, \left|1 - \alpha \lambda_{\text{max}} \right| \}, \lambda_{\text{min}} =\lambda_{\text{min}} \left(\nabla^2 \mathcal{L} (\p^*)\right),
\lambda_{\text{max}} =\lambda_{\text{max}} \left(\nabla^2 \mathcal{L} (\p^*)\right).
\]
 Furthermore,
$\alpha = \frac{2}{\lambda_{\text{min}} + \lambda_{\text{max}}}$ leads to the optimal convergence rate
\begin{equation}
  r = \frac{\kappa - 1}{\kappa + 1},\; \mbox{where}\;
  \kappa= \kappa \left(\nabla^2 \mathcal{L} (\p^*)\right) =\frac{\lambda_{\text{max}} }{\lambda_{\text{min}} }.
      \label{r_equation}
\end{equation}
\end{theorem}

This local linear convergence result can be found in \cite[Theorem 9]{Polyak1964SomeMO}, but for  quadratic functions, it is available in  \cite[Example 4.1]{10.5555/829576}.
The bound is based on optimal learning rate. In practice, a nearly optimal learning rate like (\ref{r_equation}) can be obtained through tuning. Therefore, we will consider the rate (\ref{r_equation}) as determined by the condition number when discussing GD convergence.

For minimizers with an ill-conditioned Hessian, preconditioning is a modified iteration that can be introduced to accelerate local convergence. It is an essential method in solving large linear systems of equations (i.e. quadratic optimization) and it can be formulated similarly for general optimization problems. Here, we consider a preconditioning acceleration of gradient descent  through a change of parameter as presented in  \cite{lan22}. It can also be derived as steepest descent in a non-Euclidean inner product \cite[Sec. 9.4.1]{citeulike:163662}, which is related to natural gradient descent \cite{10.1162/089976698300017746,Raiko12,Grosse15,Martens16}
in a Riemannian space that can be regarded as preconditioning for the Fisher matrix. 
Yet, another approach introduced recently in \cite{NIPS2015_f50a6c02,bernstein2024oldoptimizernewnorm} considers a local quadratic model of the loss function with the quadratic term defined by the square of a certain norm, from which a steepest descent update can be derived. This latter method is more general and leads to orthogonal matrix based updates of matrix parameters such as Shampoo \cite{pmlr-v80-gupta18a,morwani2024newperspectiveshampoospreconditioner} and MUON \cite{jordan2024muon}. 

Let $P$ be an invertible matrix and consider a change of parameters $\p = P\z$, which we call a {\em preconditioning} transformation. Writing $\mathcal{L} = \mathcal{L}(P\z)$ as a function of $\z$, GD in $\z$ is
\begin{align}
    \z_{t+1} = \z_t - \alpha \nabla_{\z}\mathcal{L}\left(P\z_t \right) = \z_t - \alpha P^{T}\nabla_{\p}\mathcal{L}\left(\p_t \right),
    \label{pgd_sgd}
\end{align}
where $\p_t=P \z_t$ and $\nabla_{\z}\mathcal{L}\left(P\z \right) = P^{T}\nabla_{\p}\mathcal{L}\left(\p \right)$. Let $\z^{*}=P^{-1}\p^*$. The corresponding optimal convergence bound is
$$\left\lVert \z_{t+1} - \z^{*} \right\rVert \leq (r+\epsilon) \left\lVert \z_t - \z^{*} \right\rVert $$ with
$r$ as determined by (\ref{r_equation}) with $\kappa$ being the condition number of the Hessian of $\mathcal{L}$ with respect to $\z$, i.e.
\[
  \kappa= \kappa \left(\nabla^{2}_{\z}\mathcal{L}\left(P\z^{*}\right)\right), \;\;{ where } \;\; \nabla^{2}_{\z}\mathcal{L}\left(P\z^{*}\right) = P^{T}\nabla^{2}_{\p}\mathcal{L}\left({\p}^{*}\right)P.
\]

If $P$ is such that $P^{T}\nabla^{2}_{\p}\mathcal{L}\left({\p}^{*}\right)P$ has a better condition number than $\nabla^{2}_{\p}\mathcal{L}\left({\p}^{*}\right)$, the local convergence rate $r$ for the GD in $\z$ (\ref{pgd_sgd}) is reduced and the convergence accelerated. Furthermore, the preconditioned iteration (\ref{pgd_sgd}) can be implemented implicitly through an iteration in $\p$ as follows.
Multiplying (\ref{pgd_sgd}) by $P$,  we obtain the following equivalent updating scheme in $\p_t=P \z_t$:
\begin{align}
    \p_{t+1} = \p_t - \alpha M \nabla_{\p}\mathcal{L}\left(\p_t\right),
    \label{pd_eqn3}
\end{align}
where $M:=PP^{T}$ is called a preconditioner. (In some literature, $(PP^{T})^{-1}$ is called a preconditioner.)
Then (\ref{pd_eqn3}) is an implicit implementation of the iteration  (\ref{pgd_sgd}), where $\z_t$ and the preconditioning transformation $\p_t=P \z_t$ are not explicitly invoked and the modified iteration (\ref{pd_eqn3}) in $\p_t$ is all that is computed. However,   (\ref{pgd_sgd}) in $\z_t$ is the underlying iteration and hence the local rate of convergence is
$\hat r = \frac{\kappa  - 1}{\kappa + 1}$, where
$ \kappa =\kappa (P^{T}\nabla^{2}_{\p}\mathcal{L}\left({\p}^{*}\right)P  )= {\lambda_{\text{max}} \left(M\nabla^2 \mathcal{L} (\p^*)\right)}/{\lambda_{\text{min}} \left(M\nabla^2 \mathcal{L} (\p^*)\right)} $.
(Note that the eigenvalues of $P^{T}\nabla^{2}_{\p}\mathcal{L}\left({\p}^{*}\right)P$ are the same as those of $M\nabla^2 \mathcal{L} (\p^*)$).  So, if $\kappa (P^{T}\nabla^{2}_{\p}\mathcal{L}\left({\p}^{*}\right)P  )$ is smaller than $\kappa (\nabla^{2}_{\p}\mathcal{L}\left({\p}^{*}\right)  )$, preconditioning accelerates the convergence.

The above theory assumes $\lambda_{\min} >0$ (i.e. strong local convexity) at a local minimum. This may not hold for a neural network that has \emph{positively scale-invariant property} (see  \cite{meng2018gsgd}). Fortunately, the above theory can be adapted to this situation with $\lambda_{\min}$ replaced by the smallest positive eigenvalue; see \cite{lan22} for details. For the rest of this paper, we assume a local minimum has positive definite Hessian.

We note that the convergence theory discussed above is limited to deterministic (or full-batch) gradient descent (GD). Recent works have explored the convergence of adaptive learning rate preconditioning with stochastic gradient descent (SGD); see   \cite{attia2023sgd,koren2022benign,faw2022power,zhou2024on}. While these analyses often establish sublinear convergence, in practice, preconditioned SGD often exhibits faster, near-linear convergence up to the level of noises. Condition number based convergence bounds have also been established in \cite{pan2024accelerated} for SGD with momentum for quadratic functions and in \cite{scott2025designingpreconditionerssgdlocal} for general preconditioned SGD. So, the Hessian conditioning-based convergence theory extends to the stochastic setting.

\section{Understanding Adaptive Learning Rate }
Accelerated gradient descent algorithms like AdaGrad, RMSProp, and Adams are widely used in deep learning. They are typically considered adaptive learning rate methods but have also been recognized as a preconditioned iteration (\ref{pd_eqn3}) with a certain preconditioner $M$; see \cite{murphy2013machine}. However, whether and how these choices of $M$ as obtained from the first order derivatives lead to improved conditioning of the Hessian is unclear.
Here, we analyze these preconditioner to show how they improve the condition number of the Hessian and thereby accelerate convergence.

\subsection{Diagonal Preconditioner}
It is well known that one way for a matrix to become ill-conditioned is that its rows (or columns) have unbalanced scaling in norm. For example, if $a_i^T$ denotes the $i$-th row of a matrix $A\in \mathbb{R}^{m\times n }$ and if the rows are orthogonal to each other, then
\[
\kappa(A) =  \kappa(D_A Q) =  \kappa(D_A ) = \frac{\max_i \|a_i\|}{\min_i \|a_i\|}
\]
where $D_A ={\rm diag}\{\|a_1\|, \cdots, \|a_m\|\}$ and $Q$ has orthonormal rows (i.e. $Q Q^T=I$). If the rows are not mutually orthogonal, then the condition number also depends on how close they are to being linearly dependent.  The ill-conditioning due to unbalanced scaling can be fixed by simply scaling its rows (or columns) to have the same norm, i.e. a diagonal preconditioner, as shown in the following theorem.

\begin{theorem}  \label{sluis_thm}
(\cite{van1969condition})
 Let $A\in \mathbb{R}^{m\times n }$ have  full column rank and
 $D_A ={\rm diag}\{ \|a_1\|, \cdots,  \|a_m\|\}$,
 where $a_i^T$ is the $i$-th row of $A$. We have
\begin{equation*}
 \kappa(D_A^{-1} A)  \le \sqrt{m} \min_{D \mbox{ is diagonal}} \kappa (D A).
 \label{sluis_ori}
 \end{equation*}
\end{theorem}

Note that $D_A^{-1} A$ amounts to scaling all rows of $A$ to have norm 1. This results in optimal condition number within a factor of $\sqrt{m}$ among all possible scalings. Clearly, scaling all rows of $A$ to have the same norm $c$ leads to  $c D_A^{-1} A$, which has the same condition number as  $D_A^{-1} A$. 
The above result holds for column scaling of $A$ (i.e. right multiplying $A$ by a diagonal matrix) as well; namely, scaling the columns of $A$ to have norm 1 achieves, within a factor of $\sqrt{n}$, optimal condition number among all possible column scalings. It can be proved by simply applying the above theorem to $A^T$ and $(DA)^T$.

In general, we are only interested in the magnitude of a condition number. Then, the following corollary shows that scaling all rows to have similar magnitudes also leads to improved conditioning.

\begin{corollary}  \label{sluis_cor}
 Let $A\in \mathbb{R}^{m\times n }$ have  full column rank. For any
 $D_0 ={\rm diag}\{d_1, \cdots, d_m \}$ with $d_i>0$, let $\gamma_i = \|a_i\|/d_i$,
 where $a_i^T$ is the $i$-th row of $A$. Then
\begin{equation*}
 \kappa(D_0^{-1} A)  \le \frac{\max_i \gamma_i}{\min_i \gamma_i} \sqrt{m}  \min_{D \mbox{ is diagonal}} \kappa (D A).
 \label{sluis_ori2}
 \end{equation*}
\end{corollary}
\begin{proof}
Let $G= {\rm diag}\{\gamma_1, \cdots, \gamma_n\}$. Then $G =  D_0^{-1} D_A$. So
\[
 \kappa(D_0^{-1} A) =  \kappa(G D_A^{-1} A)   \le  \kappa(G)  \kappa( D_A^{-1} A) \le
 \frac{\max_i \gamma_i}{\min_i \gamma_i} \sqrt{m}  \min_{D \mbox{ is diagonal}} \kappa (D A).
\]
\end{proof}

Note that $\gamma_i$ is the norm of the $i$-th row of $D_0^{-1} A$.
So, if after scaling with $D_0$, all $\gamma_i$ have similar magnitude, then the scaling achieves near optimal condition number.  For the convergence rate of gradient descent, since it is really the magnitude of the Hessian condition number that matters, a diagonal preconditioner with its $i$-th diagonal entry approximately proportional to the magnitude of $\|a_i\|^{-1}$ is sufficient to improve the conditioning.
Namely, we can construct a diagonal preconditioner  $M =diag\{1/m_1, 1/m_2, \cdots,  1/m_n \} $, where $m_i$ is approximately proportional to the  magnitude of the $i$-th row of  $\nabla^2 \Lcal  (\p^*)$.

\subsection{Adaptive Learning Rate} We now discuss how to estimate $m_i$ for a Hessian. It turns out that the gradient can provide some magnitude information on the rows of the Hessian as follows.
If $\p^*$ is a local minimum, then $\nabla\mathcal{L}(\p^*)=0$ and it follows from the Taylor expansion of $\nabla\mathcal{L}(\p_t)$ at $\p^*$ that
\[
\nabla\mathcal{L}(\p_t)= \nabla^2\mathcal{L}(\p^*) (\p_t-\p^*)+O(\lVert\p_t-\p^*\rVert^2).
\]
If we write $\mathbf{g}=\nabla\mathcal{L}(\p_t)$ and its $i$-th entry  $ {g}_i=\frac{\partial \mathcal{L}}{\partial p_i} (\p_t)$, and if
 $\mathbf{h_i}^T$ denotes the $i^{th}$ row of $\nabla^2\mathcal{L}(\p^*)$, then
\begin{eqnarray*}
{g_i}& = & \mathbf{h_i}^T(\p_t-\p^*)+O(\lVert\p_t-\p^*\rVert^2)\\
&=&\|\mathbf{h_i}\| \,\|\p_t-\p^*\| \cos \angle (\mathbf{h_i}, \p_t-\p^*) +O(\lVert\p-\p^*\rVert^2)
\end{eqnarray*}
Therefore,  barring the situation that $\mathbf{h_i}$  is nearly orthogonal to $\p_t-\p^*$ (i.e. $\cos \angle (\mathbf{h_i}, \p_t-\p^*) \approx 0$), $|g_i|$ is proportional to $\|h_i\|$ {\em in magnitude}. Namely, if we assume $\cos \angle (\mathbf{h_i}, \p_t-\p^*) >0.1$, then, for all $i$, $\frac{|g_i|}{\|h_i\|}$ is approximately within  $[0.1\|\p_t-\p^*\|, \|\p_t-\p^*\|]$. Nevertheless, the near orthogonality may occur at an iteration $t$ resulting in a much smaller $|g_i|$; but it is highly unlikely to occur over many consecutive iterations with different iterates $\p_t$.

Thus, let $m_i $ be an average of $|g_i|$ over several iterations. Since the averaging will reduce the effect of occasional occurrence of very small $ \cos \angle (\mathbf{h_i}, \p_t-\p^*)$, $m_i $ should approximately have a magnitude proportional to that of $\|h_i\|$. Using such $m_i$ to construct the preconditioner $M$ as above, the rows of the preconditioned matrix $M\nabla^2\mathcal{L}(\p^*)$ have  similar magnitudes, which by Corollary  \ref{sluis_cor} leads to improved conditioning. Indeed, different adaptive learning rate methods amount to computing different averages $m_i$ as described below.

{\bf AdaGrad} of \cite{duchi2011adaptive} computes the gradient $\mathbf{g}_t=\nabla\mathcal{L}(\p_t)$ at step $t$ and accumulates the sum of square of all the previous gradients and updates as follows:
\begin{eqnarray*}
\rb_{t+1} &\leftarrow&  \rb_{t}+\mathbf{g}_t^2 \\
\p_{t+1} &\leftarrow& \p_{t}-\alpha M_{t} \mathbf{g}_t,\;\; \mbox{ where }\;\; M_{t}={\diag} \{1/\sqrt{\rb_{t}}\}
\end{eqnarray*}
Clearly $\rb_{t}=\sum_{i=1}^t \mathbf{g}_i^2$ and $\sqrt{\rb_{t}} = \sqrt{t} \overline{\mathbf{g}}$ where $\overline{\mathbf{g}}= \sqrt{ \frac{1}{t} \sum_{i=1}^t \mathbf{g}_i^2 }$ is an equal weighted average of $|\mathbf{g}_i |$. Thus AdaGrad is a preconditioning method where the preconditioner is based on average magnitude of the gradient $ \overline{\mathbf{g}}$ with a uniform scaling $\sqrt{t}$. The scaling on the entire matrix by  $\sqrt{t}$ does not affect the condition number and hence the preconditioning property but effectively reduces $\alpha$ by $ \frac{1}{\sqrt{t}} $.

{\bf RMSProp} of \cite{RMSProp} replaces the updating formula for $\rb$ in AdaGrad by  using a running average of $\mathbf{g}_t^2$:
\begin{equation}\label{rmsprop}
\rb_{t+1}\leftarrow \rho \rb_{t}+(1-\rho)\mathbf{g}_t^2, \hspace{0.3cm}0<\rho<1.
\end{equation}
If we initially set $\rb_{1} \leftarrow \mathbf{g}_1^2$ for $t=1$, then it can be checked that at step $t>1$,
\begin{equation}\label{rmspropr}
\rb_{t}= \rho^{t-1} \mathbf{g}_1^2 + (1-\rho^{t-1}) \frac{ \sum_{i=2}^t \rho^{t-i} \mathbf{g}_i^2}{\sum_{i=2}^t \rho^{t-i}}.
\end{equation}
Therefore, $\rb_{t}$ is a weighted average of $\mathbf{g}_i^2$ with larger weights placed on later iterates.  So RMSProp is a preconditioning method where a weighted average $\overline{\mathbf{g}}= \sqrt{\rb_t }$  is used to approximate average magnitude of the gradient. One advantage of using the running average is that it uses more the recent gradients and should get a better approximation to the row-wise magnitudes of the Hessian at iteration $t$.

{\bf Adam } of \cite{2015-kingma} combines RMSProp (\ref{rmsprop}) with a momentum modified gradient that is updated as a moving average.
\begin{equation}\label{adam}
\mb_{t+1}\leftarrow \hat\rho \mb_{t}+(1-\hat\rho)\mathbf{g}_t, \hspace{0.3cm}0<\hat\rho<1.
\end{equation}
If we initially set $\rb_{1} \leftarrow \mathbf{g}_1^2$ and   $\mb_{1} \leftarrow \mathbf{g}_1$, then $\rb_{t}$ at step $t$ can be written as in (\ref{rmspropr}) with a similar formula for $\mb_{t}$. In that case, they are weighted averages of  $\mathbf{g}_i^2$ and $\mathbf{g}_i$ respectively. In this way, Adam is a preconditioning method combined with a scaled momentum. Noting that the momentum method is a generalization of the conjugate gradient method up to a scaling (see \cite{lecun-98x,Bao10081059} for example), Adam can be viewed as a generalization of preconditioned conjugate gradient  (PCG) \cite[Sec. 9.2]{10.5555/829576}. In particular, its convergence behavior follows a similar theory for PCG; namely, the momentum method with optimal momentum converges at an improved rate of $  r = \frac{\sqrt{\kappa }- 1}{\sqrt{\kappa}+ 1}$
\cite[Theorem 9]{Polyak1964SomeMO}, and by combining the preconditioning, Adam further improves convergence by reducing $\kappa$.

We note that in the original Adam of \cite{2015-kingma}, $\rb$ and $\mb$ are initially set to 0, i.e. for $t=0$, $\rb_{0} \leftarrow 0$ and   $\mb_{0} \leftarrow 0$. Then correspondingly,  (\ref{rmspropr}) becomes
$\rb_{t}=  (1-\rho^{t}) \frac{ \sum_{i=1}^t \rho^{t-i} \mathbf{g}_i^2}{\sum_{i=1}^t \rho^{t-i}}$ and similarly
$\mb_{t}=  (1-\hat\rho^{t}) \frac{ \sum_{i=1}^t \hat\rho^{t-i} \mathbf{g}_i}{\sum_{i=1}^t \hat\rho^{t-i}}$. Then $\rb_{t}$ and $\mb_{t}$ are not averages of $\mathbf{g}_i^2$ and $\mathbf{g}_i$ respectively but differ by the factors $1-\rho^{t}$ and $1-\hat\rho^{t}$ respectively called bias.  The original Adam employs a bias correction by dividing the factors $1-\rho^{t}$ and $1-\hat\rho^{t}$ to turn them into a weighted average, although, if the initialization starts with $t=1$ as  $\rb_{1} \leftarrow \mathbf{g}_1^2$ and   $\mb_{1} \leftarrow \mathbf{g}_1$ as presented here, the bias correction is not needed.

We emphasize that the key point of our analysis is that the average magnitude of each component of the gradient $\nabla\mathcal{L}(\p_t)$ can capture, up to an unknown but common factor, the corresponding row-wise magnitude information of the Hessian, but an individual gradient may not. On the other hand, there have been several works that exploit the direct relation between the gradients and the Hessian that might exist in some special situations. For example, for the cross-entropy loss, the natural gradient \cite{10.1162/089976698300017746} is a preconditioning method based on the Fisher information matrix, which is equal to the expectation of the product of the gradient with its transpose. For the mean squares error loss, the Gauss-Newton method \cite[Sec. 10.3]{GVK502988711} (or generalized Gauss-Newton \cite{JMLR:v21:17-678}) approximates the Hessian by terms essentially involving products of the gradient with its transpose. \cite{lecun-98x} suggests using only the diagonal part of these gradient products as preconditioners (i.e. the diagonal matrix of the square of the gradient), including using their moving average over several iterations. This has been suggested as one way of understanding the adaptive learning rate methods in \cite{JMLR:v21:17-678}. However, although these preconditioners resemble the adaptive learning rate methods, a critical difference is that they use {\em squared} gradients intended as an approximation of the Fisher matrix or Hessian, while the adaptive learning rate methods use the {\em square root} of the average {\em squared} gradients. Our analysis shows that this square root is essential as an estimate of the magnitudes of the rows of the Hessian, and distinguishes the corresponding preconditioners from those derived as approximations of the Hessian matrix.

\section{Regularization with Preconditioning}
Another difficulty related to adaptive learning rate arises when we have to combine them with a regularization method. Consider for example the $L_2$-regularization of a simple gradient descent, which is equivalent to weight decay. However, it is observed in \cite{loshchilov2018decoupled} that this equivalency does not  hold when an adaptive learning rate method is used. Specifically, \cite{loshchilov2018decoupled} considers a preconditioned optimizer $\p_{t+1} \leftarrow \p_t-\alpha M_t \nabla\mathcal{L}(\p_t)$ with $M_t \ne I$. It is shown in \cite{loshchilov2018decoupled} that there exists no $\lambda$ such that applying the optimizer to an $L_2$-regularized loss $ \mathcal{L}^{reg} (\p) :=\mathcal{L}(\p) +\lambda \|\p\|_2^2$ is equivalent to applying the optimizer to $\mathcal{L}(\p)$ with weight decay, i.e.
\[
\p_{t+1} \leftarrow \p_t-\alpha M_t \nabla\mathcal{L}(\p_t) -\epsilon \p_t.
\]
It is further shown that, if $M_t = {\rm diag}(1/\rb )$ for some $\rb>0$, the optimizer with weight decay is equivalent to a scale-adjusted regularization with
$ \mathcal{L}^{sreg} (\p) :=\mathcal{L}(\p) +\lambda \|\p\sqrt{\rb}\|_2^2$.  For the Adam optimizer, \cite{loshchilov2018decoupled} suggests that Adam with weight decay performs better, which they call AdamW. The intuition is that applying Adam to the $L_2$-regularization loss $ \mathcal{L}^{reg} (\p)$  results in weights with large gradient amplitudes being
regularized less than what they would be when using weight decay.

Here, we analyze these two different ways of implementing regularization in the framework of preconditioning to show that the difference lies in regularizing different parameters. In particular, AdamW amounts to selecting underlying intrinsic parameters under the preconditioning transformation for regularization. As discussed in Section \ref{pgd_section}, an accelerated GD optimizer $\p_{t+1} = \p_t-\alpha M \nabla\mathcal{L}(\p_t)$ is an  implicit implementation of the underlying GD in $\z$ under the transformation $\p = P\z$:
\[\z_{t+1} = \z_t - \alpha \nabla_{\z}\mathcal{L}\left(P\z_t \right)
\]
for the loss function $\mathcal{L} = \mathcal{L}(P\z)$, where $M=PP^T$. From this point of view, there are clearly two different ways to regularize the loss. Since the loss is considered a function of $\z$ now, it it natural to regularize with respect to $\z$ by using the loss:
\begin{equation}\label{regz}
 \mathcal{L}^{reg}_\z (\p) :=\mathcal{L}(P\z) +\lambda \|\z\|_2^2
\end{equation}
Of course, we may also regularize with respect to the original parameter $\p$:
\begin{equation}\label{regz2}
 \mathcal{L}^{reg}_\p (\p) :=\mathcal{L}(\p) +\lambda \|\p\|_2^2
\end{equation}
The two ways of regularization lead to two different algorithms with the corresponding Hessian matrices as follows.

\begin{theorem}  \label{reg}
Consider the $L_2$-regularization of the loss $\mathcal{L}(\p)$ with a preconditioning transformation  $\p = P\z$ and $M=PP^T$. Then, regularizing with respect to $\z$, i.e. using $\mathcal{L}^{reg}_\z (\p)$ of (\ref{regz}) results in a weight decay iteration \begin{equation}\label{reg1iter}
 \p_{t+1} = \p_t - \alpha M \nabla_{\p}\mathcal{L}\left(\p_t\right) -2\alpha \lambda \p_t.
\end{equation}
The corresponding Hessian matrix is $\nabla_\z^2 \mathcal{L}^{reg}_\z (\p) = P^T  \nabla_{\p}^2 \mathcal{L} (\p) P +\lambda I.$

On the other hand, regularizing with respect to $\p$, i.e. using $\mathcal{L}^{reg}_\p (\p)$ of (\ref{regz2}), results in iteration
\begin{equation}\label{reg2iter}
\p_{t+1} 
= \p_t - \alpha M \nabla_{\p}\mathcal{L}\left(\p_t\right) -2 \alpha  \lambda M \p_t .
\end{equation}
The corresponding Hessian matrix is $\nabla_\z^2 \mathcal{L}^{reg}_\p (\p) = P^T \nabla_{\p}^2 \mathcal{L} (\p) P +\lambda P^T P$.
\end{theorem}
\begin{proof}
When combining the preconditioning $\p = P\z$ with the regularization in $\z$, the loss $\mathcal{L}^{reg}_\z (\p)$ has gradient $\nabla_{\z}  \mathcal{L}^{reg}_\z (P\z) = P^{T}\nabla_{\p}\mathcal{L}\left(P\z \right) +2\lambda \z$.  Then the GD in $\z$ gives
\[
\z_{t+1} = \z_t - \alpha \left(P^{T}\nabla_{\p}\mathcal{L}\left(P\z_t \right) +2\lambda \z_t\right).
\]
Multiplying by $P$ and transforming back to the parameter $\p$, we get (\ref{reg1iter}). Taking derivative on $\nabla_{\z} \mathcal{L}^{reg}_\z (P\z)$ above leads to the Hessian.

On the other hand, when combining the preconditioning $\p = P\z$ with the regularization in $\p$, the loss $\mathcal{L}^{reg}_\p (\p)$ of (\ref{regz2}) is written as
\[
 \mathcal{L}^{reg}_\p (P\z ) = \mathcal{L}(P\z) +\lambda \|P\z \|_2^2.
\]
Then $\nabla_{\z} \mathcal{L}^{reg}_\p (P\z) = P^{T}\nabla_{\p}\mathcal{L}\left(P\z \right) +2\lambda P^TP \z$. So, the GD in $z$ gives
\[
\z_{t+1} = \z_t - \alpha \left(P^{T}\nabla_{\p}\mathcal{L}\left(P\z_t \right) +2\lambda P^TP \z_t\right).
\]
Multiplying by $P$ and transforming back to the parameter $\p$, we get (\ref{reg2iter}). Taking derivative on $\nabla_{\z} \mathcal{L}^{reg}_\p (P\z)$ leads to the Hessian.
\end{proof}

It follows from the theorem that combining Adam with regularization in $\|\z\|_2^2$ results in AdamW. Indeed,  if $P$ (or $M$) is diagonal, the corresponding loss  (\ref{regz}) is the same as $ \mathcal{L}^{sreg} (\p)=\mathcal{L}(\p) +\lambda \|\p\sqrt{\rb}\|_2^2$ for AdamW because
\[
 \mathcal{L}^{reg}_\z (\p) =\mathcal{L}(P\z) +\lambda \|\z\|_2^2 = \mathcal{L}(\p) +\lambda \|P^{-1}\p \|_2^2
\]
On the other hand, regularization with $\|\p\|_2^2$ is the same as applying $M$ directly to the gradient of $\mathcal{L}^{reg}_\p (\p)$ without taking into account of the implicit transform $\p = P\z$.

When comparing these two different regularization schemes in this framework, there are two reasons favoring regularization with respect to the intrinsic parameter $\z$. First, if we assume the preconditioner $M$ is such that the preconditioned Hessian has better conditioning, then implicitly $\z$ is a better parameterization than $\p$. For example, for a diagonal preconditioner, all rows of the Hessian with respect to $\z$ have similar magnitudes and so are $\frac{\partial^2 \mathcal{L}}{\partial z_i^2}$. Namely, the loss as a function of each parameter  $\z_i$ has similar curvature.
This justifies regularizing with $\|\z\|_2^2$, which places equal weights on all the components $\z_i$.

Furthermore, the Hessian matrices after the two regularization schemes  have very different conditioning. Regularizing $\z$ produces the Hessian $\nabla_\z^2 \mathcal{L}^{reg}_\z (\p) = P^T  \nabla_{\p}^2 \mathcal{L} (\p) P +\lambda I$ that further improves conditioning of $P^T  \nabla_{\p}^2 \mathcal{L} (\p) P$  by adding $\lambda I$. On the other hand, the regularization with $\p$ produces the Hessian $\nabla_\z^2 \mathcal{L}^{reg}_\p (\p) = P^T \nabla_{\p}^2 \mathcal{L} (\p) P +\lambda P^T P$ that may worsen the conditioning by adding $\lambda P^T P $ because $P^TP$, as a preconditioner, is expected to be an ill-conditioned matrix. Viewed another way,  $P$ is constructed to improve the conditioning of $\nabla_{\p}^2 \mathcal{L} (\p)$ but not $ \nabla_{\p}^2 \mathcal{L} (\p)   +\lambda I$.

{\bf Other Regularization Methods.}
The discrepancy between the two ways of $L_2$-regularization when combined with preconditioning also exists for other regularization methods. We state a corresponding result for $L_1$-regularization as the following theorem.

\begin{theorem}  \label{regL1}
Consider the $L_1$-regularization of the loss $\mathcal{L}(\p)$ with a preconditioning transformation  $\p = P\z$ and $M=P^2$, where $P$ is a diagonal matrix with positive diagonal entries. Then, regularizing with respect to $\z$, i.e. using $\mathcal{L}^{reg1}_\z (\p):= \mathcal{L}(P\z) +\lambda \|\z\|_1$, results in iteration
\begin{equation}\label{l1reg}
 \p_{t+1} = \p_t - \alpha M \nabla_{\p}\mathcal{L}\left(\p_t\right) -\alpha \lambda M^{1/2} \sgn (\p_t),
\end{equation}
where we assume all entries of $\p_t$ are nonzero.
On the other hand, regularizing with respect to $\p$, i.e. using $\mathcal{L}^{reg1}_\p (\p):=\mathcal{L}(\p) +\lambda \|\p\|_1$, results in iteration
\[
\p_{t+1} 
= \p_t - \alpha M  \nabla_{\p}\mathcal{L}\left(\p_t\right) - \alpha\lambda M \sgn (\p_t).
\]
\end{theorem}
\begin{proof}
For the regularization in $\z$, if all entries of $z$ are nonzero, the loss $\mathcal{L}^{reg1}_\z (\p)$ has gradient $\nabla_{\z}  \mathcal{L}^{reg1}_\z (P\z) = P^{T}\nabla_{\p}\mathcal{L}\left(P\z \right) +\lambda \sgn (\z)$.  Then the GD in $\z$ with $\mathcal{L}^{reg1}_z (\p)$ gives 
\[
\z_{t+1} = \z_t - \alpha \left(P^{T}\nabla_{\p}\mathcal{L}\left(P\z_t \right) +\lambda  \sgn( \z_t)\right).
\]
Multiplying by $P$ and noting that, since $P$ is diagonal with positive entries, $\sgn( \z_t)=  \sgn( P\z_t) = \sgn( \p_t)$, we get
\[
\p_{t+1} = \p_t - \alpha PP^{T}\nabla_{\p}\mathcal{L}\left(\p_t \right) - \alpha \lambda P \sgn( \z_t)
= \p_t - \alpha M\nabla_{\p}\mathcal{L}\left(\p_t \right) - \alpha \lambda  M^{1/2} \sgn( \p_t).
\]
The case of  regularizing with $\p$ is proved similarly.
\end{proof}

We observe that, with $L_1$-regularization in $\z$, we have a subtraction of $ M^{1/2} \sgn (\p_t)$ that is half way between  $\sgn (\p_t)$ of the non-preconditioned regularization gradient and $ M \sgn (\p_t)$ of the preconditioned regularization gradient in $\p$. Namely, it can no longer be interpreted as simply adding the gradient of the regularization term $\|\z\|_1$ to the preconditioned gradient descent as in weight decay.

Finally, there have been significant recent interests in a regularization scheme called gradient regularization where a square 2-norm of the gradient is used as the regularization term; see  \cite{barrett2020implicit,karakida2023understanding,smith2021on}.
A similar regularization using a non-squared p-norm
has also been studied in \cite{zhao2022penalizing}.
We note that only SGD optimizer has been implemented for these regularization methods. How to combine preconditioning (or adaptive learning rate) with such non-conventional regularization schemes would be an interesting future work.

\section{Normalization Methods as Preconditioning}
Normalization techniques  such as the input data normalization/standardization, batch normalization, and layer normalization can significantly accelerate training. They involve modifying the model architecture (e.g. inputs) and are {\em not} part of training algorithms. We will show that these methods benefit training by improving Hessian condition numbers. Thus, preconditioning here is not in the sense of implicit parameter transformation in GD as discussed in Section \ref{pgd_section}, but rather it directly change the model for a better optimization problem. Nevertheless, they may also be implemented as parameter preconditioning transformation as shown in recent work \cite{lan22} for batch normalization.

We consider the setting of training a fully connected neural network.
Let $y=f(x, \p): x\in \R^n \rightarrow \R^m$ be a neural network whose $\ell$-th hidden layer is defined by
\begin{equation}\label{layer}
h^{(\ell)} = g\left(W^{(\ell)}h^{(\ell-1)} + b^{(\ell)}\right)\in \mathbb{R}^{n_{\ell}};\;\; h^{(0)} = x
\end{equation}
where $W^{(\ell)} \in \R^{n_{\ell} \times n_{\ell-1 }}$ and $b^{(\ell)} \in \R^{n_\ell}$ are trainable parameters, $\p$ is the vector of all trainable parameters (i.e. all $W^{(\ell)}, b^{(\ell)}$), and 
$g(s)$ is an activation function. Let  $a^{(\ell)}:= W^{(\ell)}h^{(\ell-1)} + b^{(\ell)}$ and denote its $i$-th entry as $a^{(\ell)} (i)$. (We use this notation rather than the more conventional notation  $a_i^{(\ell)}$ because we will later use $a_i^{(\ell)}$  to denote the activation associated with the $i$-th data point.)
We have $a^{(\ell)} (i) = w_{i}^{(\ell)^T}h^{(\ell-1)} + b_{i}^{(\ell)} \in \mathbb{R}$, where $w_{i}^{(\ell)^T} \in \mathbb{R}^{1 \times n_{\ell-1}}$ and $b_{i}^{(\ell)}$ are the respective $i$th row and entry of $W^{(\ell)}$ and $b^{(\ell)}$. We also denote the $i$th entry of $h^{(\ell)}$ by  $h^{(\ell)} (i)$.  Then $h^{(\ell)} (i) = g\left(a^{(\ell)} (i) \right)$.

Given a labeled dataset $\{(x_i, y_i)\}_{i=1}^N \subset \R^m \times \R^n$,
let $L (\hat y, y)$ be some loss function for the predicted output $\hat y:= f(x, \p)$. We minimize the mean loss:
\begin{equation}\label{loss}
 \LL (\p) := \frac{1}{N}\sum_{i=1}^N L\left(f(x_i, \p), y_i\right)
\end{equation}
We will consider the Hessian of $\LL(\p)$ with respect to the weight/bias for one activation $a^{(\ell)} (i) = w_{i}^{(\ell)^T}h^{(\ell-1)} + b_{i}^{(\ell)}$.

For some fixed $\ell$ and $i$,  let
\begin{equation}
\widehat{w} = \begin{bmatrix}
       b_{i}^{(\ell)}  \\
        w_{i}^{(\ell)}
      \end{bmatrix},
      \;\;
\widehat{h} = \begin{bmatrix}
       1  \\
        h^{(\ell-1)}
      \end{bmatrix}
\mbox{ and }\; a^{(\ell)} (i) = \widehat{w}^{T}\widehat{h}.
    \label{eq:ai}
\end{equation}
Then, the output $\hat{y}= f(x,\p)$ and hence the loss $L=L (\hat y, y)$ is a function of $\widehat{w}$ through   $a^{(\ell)} (i) = \widehat{w}^{T}\widehat{h}$ only. We write this function as  $L= L_{i}^{(\ell)}\left(a^{(\ell)} (i) \right) = L_{i}^{(\ell)}\left( \widehat{w}^{T}\widehat{h} \right)$. For notational simplicity, we drop $\ell$ and $i$ to write the function as $L= L \left(a^{(\ell)} (i) \right) = L \left( \widehat{w}^{T}\widehat{h} \right)$ but note that $L \left(a_{i}^{(1)} \right)$ is not the same as $L (\hat y, y)$ even though they have the same values and we use the same letter $L$ in notation. Then the Hessian of the mean loss  with respect to  $\widehat{w}$ has a simple structure as follows.

\begin{theorem} (\cite{lan22})
\label{grad_hessian_theorem}
Let $ \{x_1, x_2, \hdots, x_N \}$ be the data inputs  to a neural network (\ref{layer}) and
let  $\{h_{1}^{(\ell-1)}, h_{2}^{(\ell-1)}, \hdots, h_{N}^{(\ell-1)} \}$ be the corresponding hidden variables $h^{(\ell-1)}$. Let $\widehat{h}_j=\begin{bmatrix}
       1  \\
        h_j^{(\ell-1)}
      \end{bmatrix}
 $ and $\widehat{w}$ as defined in (\ref{eq:ai}),
  and let $\mathcal{L} =\mathcal{L} (\widehat{w}) := \frac{1}{N}\sum_{j = 1}^{N}L\left(\widehat{w}^T\widehat{h}_j \right)$ be the mean loss. Then, its Hessian with respect to $\widehat{w}$     is
\begin{equation}
    \nabla^2_{\widehat{w}} \mathcal{L} (\widehat{w}) =  {H}_e S {H}_e^T  \label{hessian_eqn1} \end{equation}
    \text{where }
    \begin{equation} {H}_e = \begin{bmatrix}
        e^T \\
       H
      \end{bmatrix}
=  \begin{bmatrix}
      1 & \cdots & 1 \\
      h_{1}^{(\ell-1)}  & \cdots & h_{N}^{(\ell-1)}
    \end{bmatrix},
 \;\;\;
      S=\frac{1}{N}\begin{bmatrix}
       L''\left(\widehat{w}^T\widehat{h}_1\right) & & \\
        &\ddots &\\
        & &L''\left(\widehat{w}^T\widehat{h}_N\right)
      \end{bmatrix},
    \label{inputhessian_eqn1part2}
\end{equation}
$e=[1, \cdots, 1]^T$, and ${H} =[h_{1}^{(\ell-1)}, \cdots, h_{N}^{(\ell-1)}]$.
\end{theorem}

We call $H$ the hidden variable matrix and $H_e$ the extended hidden variable matrix for layer $\ell-1$.

We illustrate the above result with the linear regression and the logistic regression.
Consider the linear regression model $ \hat{y}:=w^T x + b \in \mathbb{R}$ and  $L(\hat{y}, y) =\frac{1}{2} (\hat{y}-y)^2$. By writing $\widehat{w}^{T} = \left[b, w^T \right]$,  the Hessian of the mean square loss is $\nabla^2_{\widehat{w}} \mathcal{L} (\widehat{w}) = \frac{1}{N} {X}_e {X}_e^{T}$
where
\begin{equation} {X}_e = \begin{bmatrix}
        e^T \\
       X
      \end{bmatrix} \in \mathbb{R}^{ (m+1) \times N}
      \text{ and }
      {X} =[x_1, \cdots, x_N]
    \label{inputhessian_eqn1part3}
\end{equation}
This is a special case of (\ref{hessian_eqn1}) with $\ell =1$.

Similarly, for the logistic regression model  $ \hat{y}:=\sigma(w^T x + b) \in \mathbb{R}$ with
the cross-entropy loss $L(\hat{y}, y) = -y {\log\hat{y}}-(1-y) {\log(1-\hat{y})}$, where $\sigma(s)=\frac{1}{1+e^{-s}}$ is the
logistic sigmoid function, we have
\[
\nabla^2_{\hat w} {\cal L}
= {X}_eS{X}_e^{T},\;\;
S=\frac{1}{N} \diag\{\hat{y_i}(1-\hat{y_i})\},
\]
where $\hat{y_i}= \sigma(w^T x_i + b)$. Again, it is   a special case of (\ref{hessian_eqn1}) with $\ell =1$. So Theorem \ref{grad_hessian_theorem} is a generalization of the Hessian formulas for the  linear regression and the logistic regression.

\subsection{Input Data Normalization}
Input data normalization is to transform the input data $\{x_i\}$ so that all input features (the components of $x$) follow a similar distribution. There are two common ways of normalization: one is to center the data points and then normalize by the standard deviation, often called standardization; the other is to shift and scale each feature to be bounded between 0 and 1, often called normalization; see \cite{bishop2023learning}.
As discussed in the introduction, the normalization makes intuitive sense but how it really benefits learning is not clear. Here we show that the input data standardization/normalization improves the Hessian conditioning for the input layer parameter. Note that this improved conditioning is achieved through changing the input and therefore the model as well.

Consider the input layer (i.e. $\ell=1$). For a fixed $i$, let
\begin{equation}
\widehat{w} =
\begin{bmatrix}
       b_{i}^{(1)}  \\
        w_{i}^{(1)}
      \end{bmatrix}, \;\;\;
\widehat{x} = \begin{bmatrix}
       1  \\
        x
      \end{bmatrix}
 \in \mathbb{R}^{(n +1) \times 1},\;
    \label{eq:ai2}
\end{equation}
Then $a_{i}^{(1)} = \widehat{w}^{T}\widehat{x}.$
Applying Theorem \ref{grad_hessian_theorem} to the case $\ell=1$, the Hessian of the mean loss with respect to  $\widehat{w}$ is
\begin{equation}
    \nabla^2_{\widehat{w}} \mathcal{L} (\widehat{w}) =  {X}_e S {X}_e^T \label{inputhessian_eqn1} \end{equation}
where $X_e$ is as defined in (\ref{inputhessian_eqn1part3}).
As before, we call $X$ the data matrix and $X_e$ the extended data matrix.

(\ref{inputhessian_eqn1}) shows that the condition number of the Hessian $\nabla^2_{\hat w} {\cal L}$ is roughly proportional to the square of the condition number of the extended data matrix ${X}_e$. Indeed,  $\kappa(\nabla^2_{\widehat{w}} \mathcal{L}) \le \kappa( X_e)^2 \kappa(S)$. If the $n$ features of the input $x$ are of different scales, the rows of ${X}_e$ are of different scales, which lead to ill-conditioned $X_e$ and hence ill-conditioned Hessian $\nabla^2_{\widehat{w}} \mathcal{L} (\widehat{w})$. We now show that the input normalization improves the condition number of $X_e$.

First consider the input standardization
\[
\tilde x_j= ({x}_j -\mu)/\sigma = D_x ({x}_j -\mu)
 \]
where $\mu$ and $\sigma^2$ are the respective vector mean and variance of the dataset $\{ x_j\}$, and $D_x=\diag \{ \sigma\}^{-1}$. Using the standardized dataset $\{\tilde x_j\}$ as the input, the Hessian matrix becomes $\tilde {X}_e S \tilde{X}_e^T$, where
\[
\tilde X_e = \begin{bmatrix}
     1 & \cdots  & 1 \\
      D_x({x}_1 -\mu) & \cdots &D_x({x}_N -\mu)
      \end{bmatrix}
= \tilde D_x \begin{bmatrix}
     e^T\\
      X - \mu e^T
      \end{bmatrix}
\]
and $\tilde D_x =\diag \{1, D_x\}$.  
The centered data matrix $X - \mu e^T$ is orthogonal to the first row $e^T$ because $(X - \mu e^T) e = X e -N\mu=0$. This orthogonality improves the conditioning:
\begin{equation}\label{orthcond}
 \kappa \left(\begin{bmatrix}
     e^T\\
      X - \mu e^T
      \end{bmatrix} \right) \le    \kappa \left(\begin{bmatrix}
     e^T\\
      X
      \end{bmatrix} \right);
\end{equation}
see Theorem 4 of \cite{lan22}, where it is stated in terms of the transposes of those matrices.
The amount of the improvement in conditioning depends on how large $\mu$ is. For example, if $\mu=0$, i.e. the data is already centered, then centering does not change the data and the conditioning.
Furthermore, since $\sigma_i^2$ is the variance of the $i$-th row of $X$, the norm of the $i$-th row of $X-   \mu e^T$ is $\sqrt{N} \sigma_i$. Then, after the scaling by $\sigma_i$, the $i$-th row of  $\tilde X_e$ has norm $\sqrt{N}$. Thus all rows of $ \tilde X_e$ have the same norm $\sqrt{N}$, including the first row $e^T$. Therefore, the standardization improves the conditioning of the extended data matrix $\tilde X_e$ in two ways: orthogonality and constant row norms. We state their effects in the following theorem.

\begin{theorem}\label{thm:pcond} For the extended data matrices $\tilde X_e$ (with standardization) and $X_e $ (without standardization), we have
 \begin{equation} \label{appe1}
 \kappa (\tilde X_e ) \le \sqrt{n+1} \min_D \kappa (D X_e )
 \end{equation}
\end{theorem}
\begin{proof}
Since $\sigma_i^2$ is the variance of the $i$-th row of $X$, the norm of the $i$-th row of $X-   \mu e^T$ is $\sqrt{N} \sigma_i$. Then, the norm of each row of
$\tilde X_e = \tilde D_x  \begin{bmatrix}
     e^T\\
      X - \mu e^T
      \end{bmatrix} $
is $\sqrt{N}$. By Theorem \ref{sluis_ori},
\begin{equation}\label{eqA}
\kappa (\tilde X_e ) \le \sqrt{n+1} \min_D \kappa \left( D \begin{bmatrix}
     e^T\\
      X - \mu e^T
      \end{bmatrix} \right).
\end{equation}
For any fixed diagonal matrix $D$ in the minimization, we can assume the first diagonal entry is 1, as any scaling of the entire $D$ does not change the condition number. Write $D=diag\{1, D_0\}$ where $D_0$ is an $n\times n$ diagonal matrix. So,
\[
D \begin{bmatrix}
     e^T\\
      X - \mu e^T
      \end{bmatrix}
=  \begin{bmatrix}
     e^T\\
      D_0 X - (D_0 \mu) e^T
      \end{bmatrix}.
\]
The rows of $D_0 X - (D_0 \mu) e^T$ are orthogonal to $e$ as
$\left(D_0 X - (D_0 \mu) e^T\right)e = D_0 X e - N D_0 \mu   = 0$. With this orthogonality, we can use
(\ref{orthcond}) (i.e. \cite[Theorem 4]{lan22}) to obtain
\[
 \kappa \left(\begin{bmatrix}
     e^T\\
     D_0 X - (D_0 \mu) e^T
      \end{bmatrix} \right) \le    \kappa \left(\begin{bmatrix}
     e^T\\
      D_0 X
      \end{bmatrix} \right)
 =   \kappa \left(D \begin{bmatrix}
     e^T\\
      X
      \end{bmatrix} \right).
\]
Combining this with (\ref{eqA}), we obtain 
\[
\kappa (\tilde X_e ) \le \sqrt{n+1} \min_D \kappa \left( D \begin{bmatrix}
     e^T\\
      X
      \end{bmatrix} \right) =  \sqrt{n+1} \min_D \kappa (D X_e ).
\]
\end{proof}

Now, consider normalization such as $\tilde x_i= ({x}_i -x_{\min})/(x_{\max} - x_{\min})$, where $x_{\min}$ and $x_{\max}$ are respectively the entrywise minimum and maximum vector of $\{x_i\}$, e.g. the $j$-th entry of $x_{\min}$ is the minimum of the   $j$-th entries of  $\{x_i\}$. In this case, the   $j$-th  entries of all $\tilde x_i$  are bounded between 0 and 1 with at least one of them equal to 1. Then the 1-norm of each row   of $\tilde X_e$ is bounded between 1 and $N$, including the first row. (The 2-norm of the rows are bounded between 1 and $\sqrt{N}$.) So, all rows of $\tilde X_e$ are expected to have  norms of comparable magnitude, which by Corollary \ref{sluis_cor} should improve the conditioning.

\subsection{Batch Normalization}
Batch Normalization  (BatchNorm or BN) of \cite{Ioffe15} generalizes the idea of input normalization to hidden layers of neural network.
BN for the $\ell$th layer is a linear layer inserted between the hidden variables $h^{(\ell-1)}$ and $h^{(\ell)}$ to center and scale the variable $h^{(\ell-1)}$ to have zero mean and unit variance across the mini-batch features.  Specifically,
BN replaces the  $\ell$-th hidden layer (\ref{layer}) by
\begin{equation}
\label{post_eqn}
h^{(\ell)}= g\left(W^{(\ell)} \mathcal{B}_{\beta,\gamma}(h^{(\ell-1)}) +b^{(\ell)}\right)
\end{equation}
where
\begin{equation}\label{BNoperator}
\mathcal{B}_{\beta,\gamma}\left( h^{(\ell-1)} \right) := \gamma\frac{h^{(\ell-1)} - \mu_H}{\sigma_H} + \beta,
\end{equation}
$\gamma,\beta$ are the respective trainable re-scaling and re-centering  parameter vectors, and
\begin{equation}
\label{mu}
\mu_H := 
\frac{1}{N}\sum_{j=1}^{N} {h}_j^{(\ell-1)},
\,\;
\sigma_H^2 :=  \frac{1}{N}\sum_{j=1}^{N} ({h}_j^{(\ell-1)}-\mu_H)^2
\end{equation}
are the respective vector mean and variance of $\{h_j^{(\ell-1)}\}$.
Note that BN described above is referred to as a \textit{post-activation} version.

There is a vast literature that analyzes various aspects of BN. A review is beyond the scope of this work. Our goal is to understand the accelerated training aspect of BN through the same preconditioning framework of this section. Among the existing works on analyzing BN, \cite{santurkar2019doesbatchnormalizationhelp} also demonstrates smoothing effects of BN by proving bounds that show improved Lipschitzness of the loss function and boundedness of its Hessian. Compared with their results, our analysis more directly links the normalization to the improved Hessian conditioning. In particular, this approach can be used to derive   new preconditioning algorithms; see below.

In training  a BN network, $\mu_H$ and $\sigma_H$ are considered  functions of the parameters of previous layers and the gradient with respect to those parameters would pass through $\mu_H$ and $\sigma_H$. This significantly complicates the analysis. To this end we make two simplifying assumptions.

We first observe that a post-activation BN layer defined in (\ref{post_eqn}) can be written as
\[
h^{(\ell)} 
= g\left(\widehat{W}\mathcal{B}_{0,1}\left(h^{(\ell-1)}\right) + \widehat{b}\right)
\]
where $\widehat{W} = W^{(\ell)}\text{diag}(\gamma)$ and $\widehat{b} = W^{(\ell)}\beta + b^{(\ell)}$. Namely, the $\ell$-th layer with BN operator $\mathcal{B}_{\beta,\gamma}\left( \cdot \right)$ is equivalent to one with the transformed parameters $\widehat{W}, \widehat{b}$ but normalized with $\mathcal{B}_{0,1}\left( \cdot \right)$ (that is no re-scaling and re-centering). Namely, the representation of the $\ell$-th layer by the parameters $W^{(\ell)}, b^{(\ell)},\beta,\gamma$ is an over-parameterized version and is equivalent to one using the parameters $\widehat{W}, \widehat{b}$ only. Therefore, for our analysis, we will consider normalization with $\mathcal{B}_{0,1}\left( \cdot \right)$ only.

We also assume that the gradients are not passed through $\mu_H$ and $\sigma_H$. Namely, we assume $\mu_H$ and $\sigma_H$ are constants when computing gradients for training. Under this assumption, we rewrite (\ref{post_eqn}) as
\[
h^{(\ell)}= g\left(W^{(\ell)}\tilde h^{(\ell-1)} +b^{(\ell)}\right); \;
\tilde h^{(\ell-1)} :=\mathcal{B}_{0,1}\left( h^{(\ell-1)} \right).
\]
Consider the mean loss for the input $\{x_i\}_{i=1}^N$. Applying Theorem \ref{grad_hessian_theorem} to the above form, we obtain
\[
\nabla^2_{\widehat{w}} \mathcal{L} (\widehat{w}) =  \tilde{H}_e S \tilde{H}_e^T
\]
 where
\begin{equation} \tilde{H}_e = \begin{bmatrix}
        e^T \\
       \tilde H
      \end{bmatrix},
  \text{ and }
  \tilde {H} =[\tilde h_{1}^{(\ell-1)}, \cdots, \tilde h_{N}^{(\ell-1)}].
    \label{inputhessian_h}
\end{equation}
Namely, BN  changes the Hessian in the same way as the input data normalization.
Since $\{\tilde h^{(\ell-1)}_i\}$ is standardized, the same discussions on improved conditioning of the extended data matrix $\tilde X_e$ are valid for $\tilde{H}_e $ here and show that $\tilde{H}_e $ has an improved conditioning through orthogonality and row scaling. In particular, Theorem \ref{thm:pcond} holds, i.e.
\[
 \kappa (\tilde H_e ) \le \sqrt{n_{\ell-1}+1} \min_{D \mbox{ is diagonal}} \kappa (D H_e ).
\]
Thus,  BN accelerates training by achieving nearly optimal condition number of the Hessian under the row scaling.

BN may be considered an explicit form of preconditioning like the input normalization, where we change the network hidden variables to $\tilde h^{(\ell-1)}$ to improve the conditioning of the Hessian $\nabla^2_{\widehat{w}} \mathcal{L} (\widehat{w}) =  \tilde{H}_e S \tilde{H}_e^T$. A disadvantage of this is that it makes the network architecture dependent on the mini-batch input, which results in a discrepancy between the training network and the inference network. It is observed in \cite{wang2022understanding} that this inconsistency between training and inference of BN is the leading cause for the failure of BN in the Transformer architecture for NLP. On the other hand, preconditioning presented in Section \ref{pgd_section} is to  implicitly transform the parameters for the gradient descent without changing the model; see (\ref{pd_eqn3}). Following this idea, an implicit preconditioning approach that exploits the standardization $\{\tilde h^{(\ell-1)}_i\}$ has been developed in \cite{lan22}, called Batch Normalization Preconditioning (BNP). In this setting, BNP is based on the original network (\ref{layer}) (i.e. without any normalization) but introduces a transformation on $W^{(\ell)}$ that has the same effect as the BN transform (\ref{post_eqn}) but, as a preconditioning method, can be implemented as part of training algorithm (\ref{pd_eqn3}) without modifying the network; see \cite{lan22} for details.

\subsection{Layer Normalization}
Layer Normalization  (LayerNorm or LN) of \cite{LN} normalizes a hidden variable $h^{(\ell-1)}$ along the feature dimension and has been popular for sequential models such as RNNs and Transformer; see  \cite{LN,vaswani2017attention}. Consider the  $\ell$-th hidden layer of a feedforward network that maps $h^{(\ell-1)} \in \mathbb{R}^{n_{\ell-1}}$ to  $h^{(\ell)} \in \mathbb{R}^{n_{\ell}}$ by (\ref{layer}).  Then, for a single input $x$, LayerNorm replaces the  $\ell$-th hidden layer (\ref{layer}) by
\begin{equation}
\label{LNpost_eqn}
h^{(\ell)}= g\left(W^{(\ell)} \mathcal{LN}_{\beta,\gamma}(h^{(\ell-1)}) +b^{(\ell)}\right)
\end{equation}
where
\begin{equation}\label{LNoperator}
\mathcal{LN}_{\beta,\gamma}\left( h^{(\ell-1)} \right) := \gamma\frac{h^{(\ell-1)} - \mu e}{\sigma} + \beta,
\end{equation}
$\gamma,\beta\in \mathbb{R}^{n_{\ell-1}}$ are the respective trainable re-scaling and re-centering  parameter vectors, and
\begin{equation}
\label{mu2}
\mu= \mu \left(h^{(\ell-1)}\right) := \frac{1}{n_{\ell-1}} e^T h^{(\ell-1)},
\;\;\;
\sigma  = \sigma \left(h^{(\ell-1)}\right)
:=  \frac{1}{\sqrt{n_{\ell-1}}} \|h^{(\ell-1)} - \mu e \|
\end{equation}
are the respective (scalar) mean and standard deviation of the entries of $\{h^{(\ell-1)}\}$.
Also note that (\ref{LNpost_eqn}) is a \textit{post-activation} version of LN, which is implemented in Transformer \cite{vaswani2017attention} for example. We will analyze this version.

For the input $\{x_i\}_{i=1}^N$, let  $\{ h^{(\ell-1)}_i\}_{i=1}^N$ be the corresponding hidden unit $ h^{(\ell-1)}$ of the $(\ell-1)$st layer and let
\[
\tilde h^{(\ell-1)}_i :=\mathcal{LN}_{\beta, \gamma}\left( h^{(\ell-1)}_i \right) \;\;
\mbox{ and } \;\;
h^{(\ell)}_i= g\left(W^{(\ell)}\tilde h^{(\ell-1)}_i +b^{(\ell)}\right).
\]
Applying Theorem \ref{grad_hessian_theorem} to the above form, we obtain
\[
\nabla^2_{\widehat{w}} \mathcal{L} (\widehat{w}) =  \tilde{H}_e S \tilde{H}_e^T
\]
 where
\begin{equation} \tilde{H}_e = \begin{bmatrix}
        e^T \\
       \tilde H
      \end{bmatrix},
  \text{ and }
  \tilde {H} =[\tilde h_{1}^{(\ell-1)}, \cdots, \tilde h_{N}^{(\ell-1)}].
    \label{inputhessian_h2}
\end{equation}
This compares with the Hessian ${H}_e S {H}_e^T$ without the normalization. The following theorem shows that the effect of LayerNorm then is that all columns of  $\tilde{H}_e$ have the same or comparable norms, which improves the conditioning.

\begin{theorem} Let $\tilde{H}_e$ and $H_e$ be respectively the extended hidden variable matrices with and without LayerNorm.
\begin{enumerate}
  \item If the re-scaling and bias parameters $\beta, \gamma$   in the LayerNorm are scalar parameters, then  the norm of the $i$-th column of $\tilde{H}_e$ is $\sqrt{1+n_{\ell-1}(\gamma^2 +\beta^2)}$, independent of $i$.

  \item If the re-scaling and bias parameters $\beta, \gamma$  are vector parameters, then  the norm of the $i$-th column of $\tilde{H}_e$ is bounded between 1 and $\sqrt{1+ (\sqrt{n_{\ell-1}} \| \gamma \|  +\| \beta \|)^2} $.
\end{enumerate}
\end{theorem}
\begin{proof}
First, let $v_i = \frac{h^{(\ell-1)}_i - \mu \left(h_i^{(\ell-1)}\right) e}{\sigma \left(h_i^{(\ell-1)}\right)}$. Then
\[
\tilde h^{(\ell-1)}_i =\mathcal{LN}_{\beta, \gamma}\left( h^{(\ell-1)}_i \right)
= \gamma v_i  + \beta.
\]
Note that $e^T v_i =0$ and $\|v_i\|=\sqrt{n_{\ell-1}}$.

1. If $\beta, \gamma$ are scalar parameters, then  \[
\|\tilde h^{(\ell-1)}_i \|^2 = \| \gamma v_i  + \beta e\|^2 = \gamma^2 \|v_i\|^2 + \beta^2 \|e\|^2
=n_{\ell-1}(\gamma^2 +\beta^2).
\]
Then the norm of the $i$-th column is $\sqrt{1+ \|\tilde h^{(\ell-1)}_i \|^2 } = \sqrt{1+n_{\ell-1}(\gamma^2 +\beta^2)}$.

2. On the other hand, If $\beta, \gamma$ are vector parameters, then
\[
\|\tilde h^{(\ell-1)}_i \| \le \| \gamma v_i\|  +\| \beta \| \le \| \gamma\|_{\infty} \| v_i\|  +\| \beta \| \le \sqrt{n_{\ell-1}} \| \gamma \|  +\| \beta \|.
\]
Then the norm of the $i$-th column is bounded above by $\sqrt{1+ (\sqrt{n_{\ell-1}} \| \gamma \|  +\| \beta \|)^2 }$. It is clearly bounded below by 1.
\end{proof}

If  $\beta, \gamma$  are scalar parameters, then LN results in $\tilde{H}_e$ having equal column norms, which  achieves near optimal condition number under column scaling, regardless what $\beta, \gamma$ are. In particular, if re-scaling and re-centering parameters are not used, i.e. $\beta=1$ and $\gamma=0$, then all columns of $\tilde{H}_e$ have the norm  $\sqrt{1+n_{\ell-1}}$. If   $\beta, \gamma$  are vector parameters and $\|\gamma\|$ and $\|\beta\|$ are not too large, LN results in $\tilde{H}_e$ having column norms within a small range, which also improves conditioning.

Our analysis reveals a connection between BatchNorm and LayerNorm in that they both improve the conditioning of the extended hidden variable matrix $H_e$, one through the row scaling and the other through column scaling. The centering in BatchNorm has the additional benefit of orthogonality that further improves conditioning, while the centering in LayerNorm appears to have no direct benefits. We note that it is observed empirically in \cite{NEURIPS2019_2f4fe03d} that re-centering and re-scaling by $\beta, \gamma$ in LayerNorm does not have clear benefits. Our result confirms this as long as $\|\gamma\|$ and $\|\beta\|$ are not too large.

We note that \cite{LN} analyzes LayerNorm for a generalized linear model to show its effect on the Fisher information matrix and hence the smoothness of parameter space. While there is a connection to our approach,  the Hessian conditioning analysis appears to be more intuitive that is easy to understand.

\section{Conclusion}
We have presented a unified framework of preconditioning to analyze several essential training algorithms and techniques in deep learning. Our approach offers an intuitive way to understand a very diverse class of different methods. The new perspective also clarifies how to best combine regularization with an adaptive learning rate/preconditoning method. It may also lead to new way of implementing normalization implicitly through preconditioning.

We note that we have only discussed one particular aspect of BatchNorm and LayerNorm; namely their effects in improving Hessian condition number and hence convergence acceleration. One limitation of our approach is that BatchNorm and LayerNorm have other properties, including notably invariance under weight scaling, which benefit training in other ways such as  regularization that can not be covered by the Hessian based analysis.

\vskip 0.2in

 \small
\bibliographystyle{plain}

\begin{thebibliography}{10}

\bibitem{10.1162/089976698300017746}
Shun-Ichi Amari.
\newblock Natural gradient works efficiently in learning.
\newblock {\em Neural Comput.}, 10(2):251–276, February 1998.

\bibitem{attia2023sgd}
Amit Attia and Tomer Koren.
\newblock Sgd with adagrad stepsizes: Full adaptivity with high probability to
  unknown parameters, unbounded gradients and affine variance.
\newblock In {\em International Conference on Machine Learning}, pages
  1147--1171. PMLR, 2023.

\bibitem{LN}
Jimmy Ba, Jamie Kiros, and Geoffrey Hinton.
\newblock Layer normalization.
\newblock 07 2016.

\bibitem{barrett2020implicit}
David~GT Barrett and Benoit Dherin.
\newblock Implicit gradient regularization.
\newblock In {\em International Conference on Learning Representations}, 2021.

\bibitem{bernstein2024oldoptimizernewnorm}
Jeremy Bernstein and Laker Newhouse.
\newblock Old optimizer, new norm: An anthology, 2024.

\bibitem{bishop2023learning}
Christopher~Michael Bishop and Hugh Bishop.
\newblock {\em Deep Learning - Foundations and Concepts}.
\newblock 1 edition, 2023.

\bibitem{citeulike:163662}
Stephen Boyd and Lieven Vandenberghe.
\newblock {\em Convex Optimization}.
\newblock {Cambridge University Press}, March 2004.

\bibitem{NIPS2015_f50a6c02}
David~E Carlson, Edo Collins, Ya-Ping Hsieh, Lawrence Carin, and Volkan Cevher.
\newblock Preconditioned spectral descent for deep learning.
\newblock In C.~Cortes, N.~Lawrence, D.~Lee, M.~Sugiyama, and R.~Garnett,
  editors, {\em Advances in Neural Information Processing Systems}, volume~28.
  Curran Associates, Inc., 2015.

\bibitem{duchi2011adaptive}
John Duchi, Elad Hazan, and Yoram Singer.
\newblock Adaptive subgradient methods for online learning and stochastic
  optimization.
\newblock {\em Journal of Machine Learning Research}, 12(Jul):2121--2159, 2011.

\bibitem{faw2022power}
Matthew Faw, Isidoros Tziotis, Constantine Caramanis, Aryan Mokhtari, Sanjay
  Shakkottai, and Rachel Ward.
\newblock The power of adaptivity in sgd: Self-tuning step sizes with unbounded
  gradients and affine variance.
\newblock In {\em Conference on Learning Theory}, pages 313--355. PMLR, 2022.

\bibitem{Goodfellow-et-al-2016}
Ian Goodfellow, Yoshua Bengio, and Aaron Courville.
\newblock {\em Deep Learning}.
\newblock MIT Press, 2016.

\bibitem{Grosse15}
Roger Grosse and Ruslan Salakhudinov.
\newblock Scaling up natural gradient by sparsely factorizing the inverse
  fisher matrix.
\newblock In Francis Bach and David Blei, editors, {\em Proceedings of the 32nd
  International Conference on Machine Learning}, volume~37 of {\em Proceedings
  of Machine Learning Research}, pages 2304--2313, Lille, France, 07--09 Jul
  2015. PMLR.

\bibitem{pmlr-v80-gupta18a}
Vineet Gupta, Tomer Koren, and Yoram Singer.
\newblock Shampoo: Preconditioned stochastic tensor optimization.
\newblock In Jennifer Dy and Andreas Krause, editors, {\em Proceedings of the
  35th International Conference on Machine Learning}, volume~80 of {\em
  Proceedings of Machine Learning Research}, pages 1842--1850. PMLR, 10--15 Jul
  2018.

\bibitem{Ioffe15}
S.~Ioffe and C.~Szegedy.
\newblock Batch normalization: Accelerating deep network training by reducing
  internal covariate shift.
\newblock In {\em Proceedings of the 32nd International Conference on Machine
  Learning (ICML32)}, volume~37, Lille, France, 2015. JMLR: W\&CP.

\bibitem{jordan2024muon}
Keller Jordan, Yuchen Jin, Vlado Boza, Jiacheng You, Franz Cesista, Laker
  Newhouse, and Jeremy Bernstein.
\newblock Muon: An optimizer for hidden layers in neural networks, 2024.

\bibitem{karakida2023understanding}
Ryo Karakida, Tomoumi Takase, Tomohiro Hayase, and Kazuki Osawa.
\newblock Understanding gradient regularization in deep learning: Efficient
  finite-difference computation and implicit bias.
\newblock In {\em International Conference on Machine Learning}, pages
  15809--15827. PMLR, 2023.

\bibitem{2015-kingma}
Diederik~P. Kingma and Jimmy Ba.
\newblock Adam: A method for stochastic optimization.
\newblock In {\em ICLR (Poster)}, 2015.

\bibitem{koren2022benign}
Tomer Koren, Roi Livni, Yishay Mansour, and Uri Sherman.
\newblock Benign underfitting of stochastic gradient descent.
\newblock {\em Advances in Neural Information Processing Systems},
  35:19605--19617, 2022.

\bibitem{lan22}
Susanna Lange, Kyle Helfrich, and Qiang Ye.
\newblock Batch normalization preconditioning for neural network training.
\newblock {\em Journal of Machine Learning Research}, 23(72):1--41, 2022.

\bibitem{lecun-98x}
Yann {Le Cun}, L\'{e}on Bottou, Genevieve~B. Orr, and Klaus-Robert
  M{\"{u}}ller.
\newblock Efficient backprop.
\newblock In {\em Neural Networks, Tricks of the Trade}, Lecture Notes in
  Computer Science LNCS~1524. Springer Verlag, 1998.

\bibitem{loshchilov2018decoupled}
Ilya Loshchilov and Frank Hutter.
\newblock Decoupled weight decay regularization.
\newblock In {\em International Conference on Learning Representations}, 2019.

\bibitem{JMLR:v21:17-678}
James Martens.
\newblock New insights and perspectives on the natural gradient method.
\newblock {\em Journal of Machine Learning Research}, 21(146):1--76, 2020.

\bibitem{Martens16}
James Martens and Roger Grosse.
\newblock Optimizing neural networks with kronecker-factored approximate
  curvature.
\newblock In {\em International conference on machine learning}, pages
  2408--2417, 2015.

\bibitem{meng2018gsgd}
Qi~Meng, Shuxin Zheng, Huishuai Zhang, Wei Chen, Zhi-Ming Ma, and Tie-Yan Liu.
\newblock G-{SGD}: Optimizing re{LU} neural networks in its positively
  scale-invariant space.
\newblock In {\em International Conference on Learning Representations}, 2019.

\bibitem{morwani2024newperspectiveshampoospreconditioner}
Depen Morwani, Itai Shapira, Nikhil Vyas, Eran Malach, Sham Kakade, and Lucas
  Janson.
\newblock A new perspective on shampoo's preconditioner, 2024.

\bibitem{murphy2013machine}
Kevin~P. Murphy.
\newblock {\em Machine learning : a probabilistic perspective}.
\newblock MIT Press, Cambridge, Mass. [u.a.], 2013.

\bibitem{GVK502988711}
{Jorge} Nocedal and {Stephen J.} Wright.
\newblock {\em Numerical optimization}.
\newblock Springer series in operations research and financial engineering.
  Springer, New York, NY, 2. ed. edition, 2006.

\bibitem{pan2024accelerated}
Rui Pan, Yuxing Liu, Xiaoyu Wang, and Tong Zhang.
\newblock Accelerated convergence of stochastic heavy ball method under
  anisotropic gradient noise.
\newblock In {\em The Twelfth International Conference on Learning
  Representations}, 2024.

\bibitem{Polyak1964SomeMO}
Boris Polyak.
\newblock Some methods of speeding up the convergence of iteration methods.
\newblock {\em Ussr Computational Mathematics and Mathematical Physics},
  4:1--17, 12 1964.

\bibitem{Raiko12}
Tapani Raiko, Harri Valpola, and Yann Lecun.
\newblock Deep learning made easier by linear transformations in perceptrons.
\newblock In Neil~D. Lawrence and Mark Girolami, editors, {\em Proceedings of
  the Fifteenth International Conference on Artificial Intelligence and
  Statistics}, volume~22 of {\em Proceedings of Machine Learning Research},
  pages 924--932, La Palma, Canary Islands, 21--23 Apr 2012. PMLR.

\bibitem{10.5555/829576}
Y.~Saad.
\newblock {\em Iterative Methods for Sparse Linear Systems}.
\newblock Society for Industrial and Applied Mathematics, USA, 2nd edition,
  2003.

\bibitem{santurkar2019doesbatchnormalizationhelp}
Shibani Santurkar, Dimitris Tsipras, Andrew Ilyas, and Aleksander Madry.
\newblock How does batch normalization help optimization?, 2019.

\bibitem{scott2025designingpreconditionerssgdlocal}
Mitchell Scott, Tianshi Xu, Ziyuan Tang, Alexandra Pichette-Emmons, Qiang Ye,
  Yousef Saad, and Yuanzhe Xi.
\newblock Designing preconditioners for sgd: Local conditioning, noise floors,
  and basin stability.
\newblock {\em arXiv preprint arXiv:2511.19716}, 2025.

\bibitem{smith2021on}
Samuel~L Smith, Benoit Dherin, David Barrett, and Soham De.
\newblock On the origin of implicit regularization in stochastic gradient
  descent.
\newblock In {\em International Conference on Learning Representations}, 2021.

\bibitem{RMSProp}
Tijmen Tieleman and Geoffrey Hinton.
\newblock Lecture 6.5-rmsprop: Divide the gradient by a running average of its
  recent magnitude.
\newblock {\em COURSERA: Neural networks for machine learning}, 4(2):26--31,
  2012.

\bibitem{van1969condition}
Abraham Van~der Sluis.
\newblock Condition numbers and equilibration of matrices.
\newblock {\em Numerische Mathematik}, 14(1):14--23, 1969.

\bibitem{vaswani2017attention}
Ashish Vaswani, Noam Shazeer, Niki Parmar, Jakob Uszkoreit, Llion Jones,
  Aidan~N Gomez, {\L}ukasz Kaiser, and Illia Polosukhin.
\newblock Attention is all you need.
\newblock {\em Advances in neural information processing systems}, 30, 2017.

\bibitem{Bao10081059}
Bao Wang and Qiang Ye.
\newblock Improving deep neural networks’ training for image classification
  with nonlinear conjugate gradient-style adaptive momentum.
\newblock {\em IEEE Transactions on Neural Networks and Learning Systems},
  35(9):12288--12300, 2024.

\bibitem{wang2022understanding}
Jiaxi Wang, Ji~Wu, and Lei Huang.
\newblock Understanding the failure of batch normalization for transformers in
  {NLP}.
\newblock In Alice~H. Oh, Alekh Agarwal, Danielle Belgrave, and Kyunghyun Cho,
  editors, {\em Advances in Neural Information Processing Systems}, 2022.

\bibitem{wu2018group}
Yuxin Wu and Kaiming He.
\newblock Group normalization.
\newblock In {\em Proceedings of the European conference on computer vision
  (ECCV)}, pages 3--19, 2018.

\bibitem{NEURIPS2019_2f4fe03d}
Jingjing Xu, Xu~Sun, Zhiyuan Zhang, Guangxiang Zhao, and Junyang Lin.
\newblock Understanding and improving layer normalization.
\newblock In {\em Advances in Neural Information Processing Systems},
  volume~32, 2019.

\bibitem{zhao2022penalizing}
Yang Zhao, Hao Zhang, and Xiuyuan Hu.
\newblock Penalizing gradient norm for efficiently improving generalization in
  deep learning.
\newblock In {\em International Conference on Machine Learning}, pages
  26982--26992. PMLR, 2022.

\bibitem{zhou2024on}
Dongruo Zhou, Jinghui Chen, Yuan Cao, Ziyan Yang, and Quanquan Gu.
\newblock On the convergence of adaptive gradient methods for nonconvex
  optimization.
\newblock {\em Transactions on Machine Learning Research}, 2024.
\newblock Featured Certification.

\end{thebibliography}

\end{document}